\documentclass[10pt,twocolumn,letterpaper]{article}

\usepackage[pagenumbers]{cvpr} 

\usepackage[table]{xcolor}

\definecolor{cvprblue}{rgb}{0.21,0.49,0.74}
\usepackage[pagebackref,breaklinks,colorlinks,citecolor=cvprblue]{hyperref}

\usepackage[utf8]{inputenc} 
\usepackage[T1]{fontenc}    
\usepackage{url}            
\usepackage{booktabs}       
\usepackage{amsfonts}       
\usepackage{nicefrac}       
\usepackage{microtype}      
\usepackage{amsmath}
\usepackage{amssymb}
\usepackage{multirow}

\usepackage[accsupp]{axessibility} 

\def\paperID{2434} 
\def\confName{CVPR}
\def\confYear{2024}

\title{Point2RBox: Combine Knowledge from Synthetic Visual Patterns for End-to-end Oriented Object Detection with Single Point Supervision\vspace{-4pt}}

\author{\textbf{Yi Yu}$^{1*}$,~\textbf{Xue Yang}$^{2}$\thanks{Equal contribution. $^{\dag}$Correspondence author. The work was partly supported by National Natural Science Foundation of China (62306069, 62222607, 72342023), Special Project on Basic Research of Frontier Leading Technology of Jiangsu Province of China (BK20192004C), China Postdoctoral Science Foundation (2023M740602), National Key R\&D Program of China (2022ZD0160100). Yi Yu is also supported by Jiangsu Funding Program for Excellent Postdoctoral Talent (2023ZB616).}\,\,,~\textbf{Qingyun Li}$^{3,2}$,~\textbf{Feipeng Da}$^{1\dag}$,~\textbf{Jifeng Dai}$^{4,2}$,~\textbf{Yu Qiao}$^2$,~\textbf{Junchi Yan}$^{5,2\dag}$\\
$^{1}$Southeast University \quad $^{2}$Shanghai AI Laboratory \quad
$^{3}$Harbin Institute of Technology \\ $^{4}$Tsinghua University \quad $^{5}$Shanghai Jiao Tong University \\
{\vspace{-3pt}\tt\small \{yuyi,dafp\}@seu.edu.cn \quad \{yangxue,qiaoyu\}@pjlab.org.cn}\\
{\vspace{-3pt}\tt\small 21b905003@stu.hit.edu.cn \quad daijifeng@tsinghua.edu.cn \quad yanjunchi@sjtu.edu.cn}\\
{\vspace{-10pt}\tt\small \url{https://github.com/yuyi1005/point2rbox-mmrotate}}}

\begin{document}
\maketitle
\begin{abstract}
\vspace{-4pt}
With the rapidly increasing demand for oriented object detection (OOD), recent research involving weakly-supervised detectors for learning rotated box (RBox) from the horizontal box (HBox) has attracted more and more attention. In this paper, we explore a more challenging yet label-efficient setting, namely single point-supervised OOD, and present our approach called Point2RBox. Specifically, we propose to leverage two principles: 1) Synthetic pattern knowledge combination: By sampling around each labeled point on the image, we spread the object feature to synthetic visual patterns with known boxes to provide the knowledge for box regression. 2) Transform self-supervision: With a transformed input image (e.g. scaled/rotated), the output RBoxes are trained to follow the same transformation so that the network can perceive the relative size/rotation between objects. The detector is further enhanced by a few devised techniques to cope with peripheral issues, e.g. the anchor/layer assignment as the size of the object is not available in our point supervision setting. To our best knowledge, Point2RBox is the first end-to-end solution for point-supervised OOD. In particular, our method uses a lightweight paradigm, yet it achieves a competitive performance among point-supervised alternatives, 41.05\%/27.62\%/80.01\% on DOTA/DIOR/HRSC datasets. 
\end{abstract}

\vspace{-2pt}
\section{Introduction}\label{sec:introduction}
\vspace{-4pt}
As a fundamental task in computer vision, object detection plays an important role, e.g. in autonomous driving \cite{feng2021deep}, aerial images \cite{Xia2018DOTA, Liu2017HRSC, Yang2018Automatic, Yang2022Arbitrary, Fu2020Rotation}, scene text \cite{Nayef2017ICDAR, Ma2018Arbitrary, Liao2018Rotation, Liu2018FOTS, Zhou2017EAST}, retail scenes \cite{goldman2019precise, pan2020dynamic}, industrial inspection \cite{Liu2020Data, Wu2022PCBNet}, and more, with the object detection results usually rendered in three ways: \textbf{1)} Horizontal bounding box (HBox); \textbf{2)} Rotated bounding box (RBox); \textbf{3)} Pixel-wise labels (Mask).

To teach the detector new concepts of visual objects, manual annotations are required. Early research is usually based on full supervision where the manual annotation is in the same manner as the desired network output. Although having achieved promising performance, full supervision in oriented object detection faces with two problems: 
\textbf{1)} RBox annotations are less available in many scenarios. In particular, a large number of datasets have already been annotated with other formats. When RBoxes are required, a possible way is to conduct labor-intensive re-annotation, e.g. DIOR-HBox \cite{li2020object} to DIOR-RBox \cite{cheng2022anchor} for aerial image (192K instances) and SKU110K-HBox \cite{goldman2019precise} to SKU110K-RBox \cite{pan2020dynamic} for retail scene (1,733K instances). \textbf{2)} RBox annotations are much more costly. The cost of each RBox is about 36.5\% higher than an HBox and 104.8\% higher than a point annotation\footnote{According to \url{https://cloud.google.com/ai-platform/data-labeling/pricing} and point annotations are 1.1-1.2x more time consuming than obtaining image-level labels \cite{bearman2016point}.}.

To mitigate the dependence on labor-intensive RBox labeling, H2RBox \cite{yang2023h2rbox} and H2RBox-v2 \cite{yu2023h2rboxv2} have explored the HBox-to-RBox setting that learns RBox detectors from HBox annotations. A more challenging task setting is then featured: Can we achieve oriented object detection under the weak supervision of point annotations? 

\begin{figure*}[t!]
\setlength{\abovecaptionskip}{1.2mm}
\centering
\includegraphics[width=1.0\linewidth]{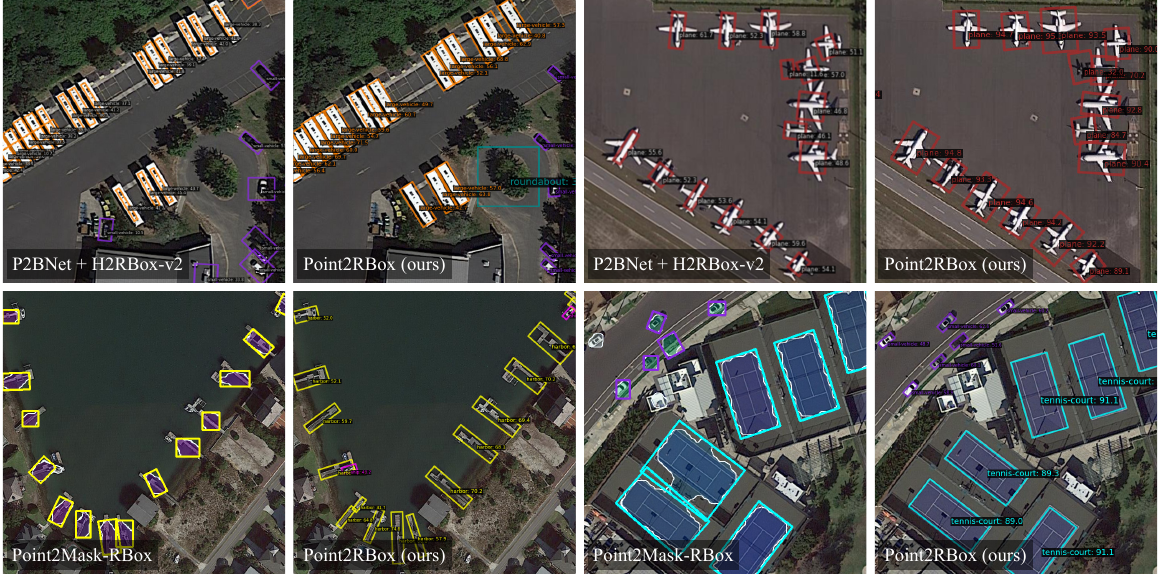}
\caption{Visual detection results based on the same ResNet50 \cite{He2016Deep} backbone. The first row compares our method (Point2RBox-SK, AP$_{50}$ = 40.27, see Table \ref{tab:exp_dota}) with Point-to-HBox-to-RBox pipeline powered by the state-of-the-art P2BNet (2022) \cite{chen2022pointtobox} and H2RBox-v2 (2023) \cite{yu2023h2rboxv2}. The second row displays the comparison with Point-to-Mask-to-RBox method Point2Mask-RBox \cite{li2023point2mask}.}
\label{fig:vis}
\vspace{-12pt}
\end{figure*}

Several point supervised detectors for HBox/Mask have been proposed: \textbf{1)} P2BNet \cite{chen2022pointtobox} samples boxes of different sizes around the labeled point and classify them to achieve Point-to-HBox; \textbf{2)} Point2Mask \cite{li2023point2mask} achieves segmentation using a single point annotation per target for training; \textbf{3)} SAM (Segment Anything Model) \cite{kirillov2023segany} produces object masks from input Point/HBox prompts. Although the above Point-to-HBox/Mask settings can be applied to Point-to-RBox, e.g. by using an additional HBox-to-RBox stage or finding the circumscribed rectangle of the mask, we show that such a solution is not perfect in both speed (7x slower converting from mask \cite{yang2023h2rbox}) and accuracy (see Fig.~\ref{fig:vis}). 

\textbf{Motivation.} Using the finger for single-point instructions is a natural way to convey the object concepts. While point supervision is gaining attention \cite{chen2022pointtobox,li2023point2mask}, end-to-end point-supervised oriented detection is still a missing part of the literature. Inspired by humans' ability to use knowledge from sketches to learn real-world objects, we intend to explore Point-to-RBox with a similar and novel idea -- using synthetic visual patterns for knowledge combination.

\textbf{What is new?} \textbf{1)} A light-weight detector Point2RBox for leaning RBox from single point annotations is proposed. It has a simple and elegant structure and is trained in a single-stage end-to-end manner. \textbf{2)} At the core of Point2RBox is novel principles for RBox regression: synthetic pattern knowledge combination and transform self-supervision. The former enables the network to estimate the size and angle of real objects through the knowledge from synthetic patterns, whereas the latter to perceive the relative size/rotation between objects. \textbf{3)} As a result, Point2RBox give a competitive performance, as is displayed in Fig.~\ref{fig:vis} and Tables \ref{tab:exp_dota}-\ref{tab:exp_hrsc}.

\textbf{Contributions.} \textbf{1)} To our best knowledge, this work is the first attempt for end-to-end single point supervised OOD, where we propose two schemes: knowledge combination and self-supervision. \textbf{2)} The training pipeline and detail implementation are elucidated, with devised techniques to cope with peripheral issues, e.g. the anchor assignment when the object size is not available. \textbf{3)} Extensive experiments demonstrate the method's capability to learn RBox regression, surpassing other alternatives in performance. 

\begin{figure*}[t]
\setlength{\abovecaptionskip}{1.2mm}
\centering
\includegraphics[width=1\linewidth]{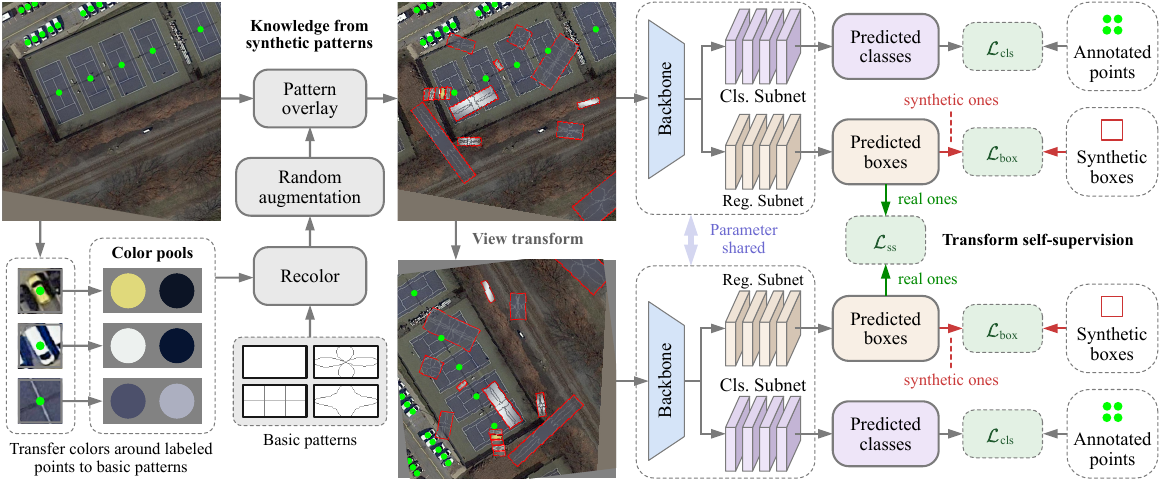}
\caption{The training flowchart, consisting of synthetic pattern knowledge combination (Sec.~\ref{sec:skc}) and transform self-supervision  (Sec.~\ref{sec:tss}). The core idea is to combine knowledge from synthetic patterns for size/angle estimation with that from annotated points for classification. The basic patterns are obtained based on two different settings (see Fig. \ref{fig:settings}).}
\label{fig:point2rbox}
\vspace{-8pt}
\end{figure*}

\section{Related Work}\label{sec:related_work}
\vspace{-4pt}
Beyond horizontal detection \cite{Zhao2019Object,liu2020deep}, oriented object detection \cite{wen2023comprehensive} has received extensive attention. Here, pproaches to oriented detection and point supervision are discussed.

\textbf{Fully-supervised oriented detection.} Representative works include anchor-based detector Rotated RetinaNet \cite{Lin2017Focal}, anchor-free detector Rotated FCOS \cite{Tian2019FCOS}, and two-stage solutions, e.g. RoI Transformer \cite{Ding2018Learning}, Oriented R-CNN \cite{Xie2021Oriented}, and ReDet \cite{Han2021Redet}. Some research enhances the detector by exploiting alignment features, e.g. R$^3$Det \cite{Yang2021R3Det} and S$^2$A-Net \cite{Han2022Align}. The angle regression may face boundary discontinuity and remedies are developed, including modulated losses \cite{Yang2019SCRDet, Qian2021RSDet} that alleviate loss jumps, angle coders \cite{Yang2020Arbitrary, Yang2021Dense, yu2023psc} that convert the angle into boundary-free coded data, and Gaussian-based losses \cite{Yang2021Rethinking, Yang2021Learning, yang2023detecting, yang2023kfiou} transforming rotated bounding boxes into Gaussian distributions. RepPoint-based methods \cite{Yang2019Reppoints, hou2022grep, li2022oriented} provide alternatives that predict a set of sample points that bounds the spatial extent of an object.

\textbf{HBox-to-RBox.} The seminal work H2RBox \cite{yang2023h2rbox} circumvents the segmentation step and achieves RBox detection directly from HBox annotation. With HBox annotations for the same object in various orientations, the geometric constraint limits the object to a few candidate angles. Supplemented with a self-supervised branch eliminating the undesired results, an HBox-to-RBox paradigm is established. An enhanced version H2RBox-v2 \cite{yu2023h2rboxv2} is proposed to leverage the reflection symmetry of objects to estimate their angle, further boosting the HBox-to-RBox performance.

Some similar studies use additional annotated data for training, which are also attractive but less general: \textbf{1)} OAOD \cite{iqbal2021leveraging} is proposed for weakly-supervised oriented object detection. But in fact, it uses HBox along with an object angle as annotation, which is just “slightly weaker” than RBox supervision. Such an annotation manner is not common, and OAOD is only verified on their self-collected ITU Firearm dataset. 
\textbf{2)} Sun et al. \cite{sun2021oriented} propose a two-stage framework: i) training detector with the annotated horizontal and vertical objects, and ii) mining the rotation objects by rotating the training image to align the oriented objects as horizontally or vertically as possible.
\textbf{3)} KCR \cite{zhu2023knowledge} combines RBox-annotated source datasets with HBox-annotated target datasets, and achieves HBox-supervised oriented detection on the target datasets via transfer learning.

\textbf{Point-to-HBox.} Several related approaches have been developed, including: \textbf{1)} P2BNet \cite{chen2022pointtobox} samples box proposals of different sizes around the labeled point and classify them to achieve point-supervised horizontal object detection. \textbf{2)} PSOD \cite{gao2022weakly} achieves point-supervised salient object detection using an edge detector and adaptive masked flood fill.

Some methods accept partial point annotations (a common setting is 80\% points and 20\% HBoxes), usually termed semi-supervision: 
\textbf{1)} Point DETR \cite{chen2021points} extends DETR \cite{carion2020detr} by adding a point encoder for point annotations. 
\textbf{2)} Group-RCNN \cite{zhang2022grouprcnn} generates a group of proposals for each point annotation.
\textbf{3)} CPR \cite{yu2022object} produces center points from coarse point annotations, relaxing the supervision signals from accurate points to freely spotted points.

These Point-to-HBox methods are potentially applicable to our Point-to-RBox task setting -- by using a subsequent HBox-to-RBox stage. In our experiment, the state-of-the-art methods P2BNet \cite{chen2022pointtobox} and H2RBox-v2 \cite{yu2023h2rboxv2} are used to build a Point-to-HBox-to-RBox baseline for comparison.

\textbf{Point-to-Mask.} Compared with point-supervised oriented detection, Point-to-Mask has been better studied. For instance, Point2Mask \cite{li2023point2mask} is proposed to achieve panoptic segmentation using only a single point annotation per target for training. SAM (Segment Anything Model) \cite{kirillov2023segany} produces object masks from input point/HBox prompts.

The Point-to-Mask pipeline is also a potential alternative for our task -- by finding the minimum circumscribed rectangle of the segmentation mask. Though the oriented bounding box can be obtained from the mask, such a complex pipeline can be less cost-efficient and perform worse \cite{yang2023h2rbox, yu2023h2rboxv2}. 

\section{Method}\label{sec:method}
\vspace{-4pt}

\begin{figure*}[t]
\setlength{\abovecaptionskip}{1.2mm}
\centering
\includegraphics[width=1\linewidth]{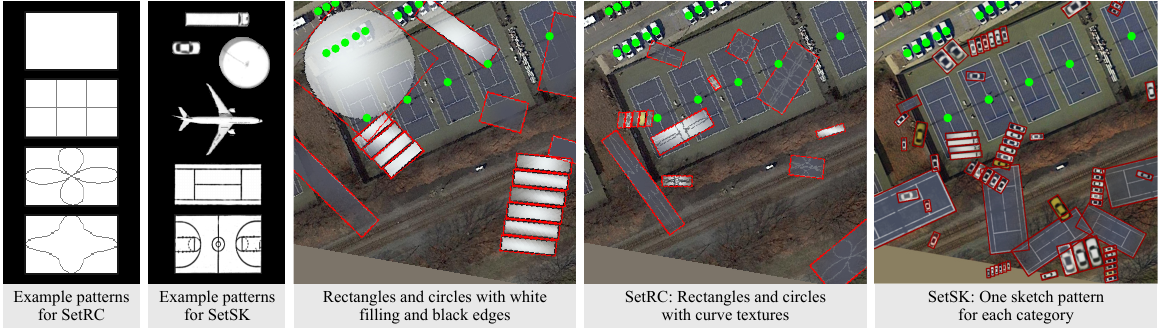}
\caption{Two settings of obtaining basic patterns (see Sec. \ref{sec:skc}) and the illustration of training images overlaid with synthetic patterns. SetRC: Rectangles and circles with curve textures. SetSK: One simple sketch pattern for each category (see Table \ref{tab:abl_pattern} for ablation).}
\label{fig:settings}
\vspace{-8pt}
\end{figure*}

An overview of the proposed Point2RBox is illustrated in Fig.~\ref{fig:point2rbox}, which consists of synthetic pattern knowledge combination (Sec.~\ref{sec:skc}) and transform self-supervision (Sec.~\ref{sec:tss}). 

On the left is the generation procedure for the knowledge combination. In this phase, synthetic patterns are generated and randomly overlaid on the input image. These patterns with known bounding boxes are subsequently used to supervise the regression during the loss calculation. 

After the image is generated, we perform a transformation on this image (randomly selected from rotate, flip, and scale). By feeding both the original view and the transformed view into the network, we obtain two sets of output RBoxes. The transform self-supervision loss is calculated between these two sets so that the output RBoxes are trained to follow the same transformation applied to the input.

In the subsequent subsections, these modules are further detailed, with the loss function described in Sec.~\ref{sec:loss}. 

\subsection{Synthetic Pattern Knowledge Combination}\label{sec:skc}
\vspace{-2pt}
In single point-supervised OOD, we do not know the exact size/angle of the labeled object. Yet we can generate a new one -- with a similar texture and known bounding box. Based on this idea, we sample around each labeled point, extract its neighbor color feature, spread them to basic patterns, and overlay them on the input image. These patterns are then used to enable the training of the regression subnet.

\textbf{Basic patterns.} \textbf{1) SetRC} \textit{(point supervision)}. It consists of rectangles and circles with white filling and black edges, sized 160$\times$160, 160$\times$80, 160$\times$40, 80$\times$80, 80$\times$40, and 80$\times$20. The above preset is used to keep the size of generated patterns in a reasonable range. \textbf{2) SetSK} \textit{(point supervision with one-shot samples)}. It consists of one sketch pattern for each category (e.g. only fifteen for DOTA-v1.0). The patterns are manually cropped from the first several training images and are adjusted to ``white surfaces and black edges''. It takes just about ten minutes to obtain all fifteen patterns for the DOTA-v1.0 dataset.

In the meanwhile, curve patterns can be overlaid on the shapes to improve the diversity (in SetRC, see Fig. \ref{fig:settings}). Two types of textures are used: \textbf{1)} One to four equally spaced lines parallel to bounding boxes. \textbf{2)} The curve defined by the following polar coordinate equation:
\begin{equation}
\setlength{\abovedisplayskip}{6pt}
\setlength{\belowdisplayskip}{6pt}
\rho  = (1-k)\left | \cos^n(2\theta) \right | +k
\end{equation}
where $n$ is a random number in $[0, 8]$, and $k$ is a random number in $[0.1, 0.6]$, both from uniform distribution. The two types of textures are adopted each with a probability of 0.5. Several example patterns are displayed in Fig. \ref{fig:settings}.

\textbf{Color sampling.} The face color $C_\text{face}$ and the edge color $C_\text{edge}$ around each labeled point are extracted as follows:
\begin{equation}
\setlength{\abovedisplayskip}{6pt}
\setlength{\belowdisplayskip}{6pt}
\left\{\begin{array}l
 C_\text{face}=\text{mean}\left ( I_0  \right ) 
 \\
 C_\text{edge}=\text{sum}\left ( dI_1 \right ) 
\end{array}\right.
\end{equation}
where $I_0$ and $I_1$ are the neighbors around a labeled point. We simply use $5\times5$ for $I_0$ and $33\times33$ for $I_1$. Here $d$ is the gradient of $I_1$ indicating the edge intensity of each pixel (the sum of $d$ is uniformed to 1).

\textbf{Recolor.} 
We use a gray-scale image as the basic pattern, which can be denoted as $P$, with its value in the range $(0, 1)$. The recolor step can be expressed as:
\begin{equation}
\setlength{\abovedisplayskip}{6pt}
\setlength{\belowdisplayskip}{6pt}
P_\text{recolor}=PC_\text{face}+\left ( 1-P \right ) C_\text{edge}
\end{equation}

This formula maps the extracted color $C_\text{face}$ and $C_\text{edge}$ to the basic pattern. By this means, the diversity of the patterns is significantly enriched, so that the trained regression subnet can better estimate the RBox of the real data. 

\textbf{Random augmentation.} The recolored patterns are first augmented with the random flip, resize, and rotation. Afterward, they are moved to a random position inside the image border. To avoid overlapping patterns, NMS (Non-Maximum Suppression) is then applied so that the IoU (Intersection over Union) between patterns is less than 0.05. 

The random resize can be formulated as:
\begin{equation}
\setlength{\abovedisplayskip}{6pt}
\setlength{\belowdisplayskip}{6pt}
\left\{\begin{array}l
s_\text{base} = \exp\left ( \sigma _\text{base}\text{randn()} \right )  \\
w = s_\text{base}\exp\left ( \sigma_w\text{randn()} \right ) w_0  \\
h = \frac{h_0}{w_0} \exp\left ( \sigma_r\text{randn()} \right )w
\end{array}\right.
\end{equation}
where $\text{randn()}$ generates random numbers from the standard normal distribution; $w_0$ and $h_0$ are the original pattern sizes; $w$ and $h$ are the resized ones; $s_\text{base}$ is a base scale for all patterns in the same image; $\sigma _\text{base}$, $\sigma_w$, and $\sigma_r$ control the variation of scale, width, and ratio, set to 0.4 by default.

For the other augmentation i.e. random flip and rotation, the probability is set to 0.5 and 1, respectively. 

In particular, to avoid the real objects being completely occluded, we use transparent blending:
\begin{equation}
\setlength{\abovedisplayskip}{6pt}
\setlength{\belowdisplayskip}{6pt}
t(x,y)=\exp\left ( -ax^2-by^2 \right ) \times t_1+t_0
\end{equation}
where $t(x,y)$ is the opacity channel of the synthetic pattern; $x$ and $y$ are coordinates in range $[-1, 1]$; $a$ and $b$ are random numbers in $[0.1, 2]$ from uniform distribution; $t_0=0.1$ and $t_1=0.9$ keep the opacity between 10\% to 100\%.

In addition, some of the patterns will be randomly selected to produce a set of tightly arranged patterns through translation, which can also be observed in Fig. \ref{fig:settings}.

Finally, these generated patterns are overlaid on the original image and their known bounding boxes are used for training, providing the knowledge for box regression.

\subsection{Transform Self-supervision}\label{sec:tss}
\vspace{-2pt}
Our training flowchart has two parameter-shared branches (see Fig. \ref{fig:point2rbox}), namely the original branch and the transformed branch (i.e. with a transformed input). The self-supervision is performed between the two branches to supervise each other. The input of the transformed branch can be set as:
\begin{equation}
\setlength{\abovedisplayskip}{6pt}
\setlength{\belowdisplayskip}{6pt}
I_\text{trs}=\mathit{transform}\left ( I_\text{ori} \right )
\end{equation}
where $I_\text{ori}$ is the input of the original branch; $I_\text{trs}$ is the input of the transformed branch; $\mathit{transform}\left ( \cdot \right )$ is a transformation randomly selected from flip, rotate, and scale. The probability of scale is set to 30\%, and the proportion between rotate and flip is set to 95:5 adopted from \cite{yu2023h2rboxv2}.

The output RBoxes of the two branches are defined as:
\begin{equation}
\setlength{\abovedisplayskip}{6pt}
\setlength{\belowdisplayskip}{6pt}
\left\{\begin{array}l
  B_\text{ori}=\left ( x_\text{ori},y_\text{ori},w_\text{ori},h_\text{ori},\theta_\text{ori} \right )  \\
  B_\text{trs}=\left ( x_\text{trs},y_\text{trs},w_\text{trs},h_\text{trs},\theta_\text{trs} \right ) 
\end{array}\right.
\label{equ:rboxdef}
\end{equation}
where $B_\text{ori}$ is the output RBoxes of original branch; $B_\text{trs}$ is those of transformed branch.

The two branches supervise each other by minimizing the loss between $B_\text{ori}$ and $B_\text{trs}$ so that the transformation between $B_\text{ori}$ and $B_\text{trs}$ is trained to follow the same transformation being applied to the input image.

It should be noted that only the output RBoxes assigned to real objects (i.e. point-annotated ones) are involved in transform self-supervision, whereas synthetic ones are directly supervised by their known bounding boxes.

\textbf{Flip.} When the input image is vertically flipped, the angle of output RBoxes is also supposed to be flipped. Therefore, the self-supervised loss can be expressed as:
\begin{equation}
\setlength{\abovedisplayskip}{6pt}
\setlength{\belowdisplayskip}{6pt}
L_\text{flp}\left ( B _\text{ori}, B_\text{trs} \right )  = smooth_{L1}  \left ( mod(\theta _\text{trs} + \theta_\text{ori} ), 0 \right ) 
\label{equ:loss_flp}
\end{equation}
where $\theta _\text{ori}$ is the output angle of original branch; $\theta _\text{trs}$ is the output angle of transformed branch; $mod(x) = (x + \pi/2 \mod \pi) - \pi/2$ limits the difference in range $\left[-\pi/2, \pi / 2\right)$.

\textbf{Rotate.} When the input image is rotated by $\mathcal{R}$, the angle of output RBoxes is supposed to be likewise rotated. Therefore, the self-supervised loss can be expressed as:
\begin{equation}
\setlength{\abovedisplayskip}{6pt}
\setlength{\belowdisplayskip}{6pt}
L_\text{rot}\left ( B _\text{ori}, B_\text{trs} \right )  = smooth_{L1}  \left ( mod(\theta _\text{trs} - \theta_\text{ori} ), \mathcal{R} \right ) 
\label{equ:loss_rot}
\end{equation}
where $\theta _\text{ori}$,  $\theta _\text{trs}$, and $mod(x)$ share the same definition as Eq. (\ref{equ:loss_flp}); $\mathcal{R}$ is the rotation angle being applied to the input image in range $\left( 0.25\pi, 0.75\pi \right)$ adopted from \cite{yu2023h2rboxv2}.

\textbf{Scale.} When the input image is scaled by $s$, the center and the size of output RBoxes should be likewise scaled. Therefore, the self-supervised loss can be expressed as:
\begin{equation}
\setlength{\abovedisplayskip}{6pt}
\setlength{\belowdisplayskip}{6pt}
L_\text{sca}\left ( B _\text{ori}, B_\text{trs} \right )  = \mathit{GIoU}  \left ( \mathit{r2h} (B _\text{ori}) \times s, \mathit{r2h}(B_\text{trs}) \right ) 
\label{equ:loss_sca}
\end{equation}
where $B _\text{ori}$ and $B _\text{trs}$ are outputs of the original and transformed branches; $\mathit{r2h}\left(\cdot\right)$ is the function to get the circumscribed HBoxes of RBoxes, $s$ is the scaling factor being applied to the input image in range $\left( 0.5, 1.5 \right)$.

\subsection{Loss Functions}\label{sec:loss}
\vspace{-2pt}
The predicted RBoxes and the corresponding ground-truths are denoted as $B_\text{pred}$ and $B_\text{gt}$.

\textbf{Loss from point supervision.} Point annotations are used to train the classification and the center of RBoxes:
\begin{equation}
\setlength{\abovedisplayskip}{6pt}
\setlength{\belowdisplayskip}{6pt}
\left\{\begin{array}l
\mathcal{L}_\text{cls} = L_\text{cls}\left ( M_\text{point}c_\text{pred}, M_\text{point}c_\text{gt}\right )
 \\
\mathcal{L}_\text{cen} = L_1\left ( M_\text{point}xy_\text{pred}, M_\text{point}xy_\text{gt}\right )
\end{array}\right.
\end{equation}
where $M_\text{point}$ is a mask to select those RBoxes that are assigned to annotated points; $c_\text{pred}$ and $xy_\text{pred}$ are the predicted classification scores and centers; $c_\text{gt}$ and $xy_\text{gt}$ are the labels and coordinates of annotated points; $L_\text{cls}$ is the classification loss defined by backbone oriented detector.

\textbf{Loss from knowledge combination.} Knowledge from synthetic patterns is combined to learn the box regression:
\begin{equation}
\setlength{\abovedisplayskip}{6pt}
\setlength{\belowdisplayskip}{6pt}
\mathcal{L}_\text{box} = L_\text{box}\left ( M_\text{box}B_\text{pred}, M_\text{box}B_\text{gt}\right )
\end{equation}
where $M_\text{box}$ is a mask to select those RBoxes that are assigned to synthetic boxes; $L_\text{box}$ is the loss function depending on the backbone oriented detector.

\textbf{Loss from transform self-supervision.} The loss is calculated between the original and transformed branches and the loss function is selected corresponding to the type of transformation being applied to the input image:
\begin{equation}
\setlength{\abovedisplayskip}{6pt}
\setlength{\belowdisplayskip}{6pt}
\mathcal{L}_\text{ss} = L_\text{flp/rot/sca}\left ( M_\text{ori}M_\text{point}B_\text{pred}, M_\text{trs}M_\text{point}B_\text{pred}\right )
\end{equation}
where $L_\text{flp/rot/sca}$ is defined by Eqs.~(\ref{equ:loss_flp}, \ref{equ:loss_rot}, \ref{equ:loss_sca}); $M_\text{ori}$ and $M_\text{trs}$ are the masks to select those RBoxes from the original branch or transformed branch. 

\textbf{Overall loss.} The overall loss of the proposed network is the weighted sum of the above losses:
\begin{equation}
\setlength{\abovedisplayskip}{6pt}
\setlength{\belowdisplayskip}{6pt}
\mathcal{L}_\text{total} = \omega_\text{cls}\mathcal{L}_\text{cls} + \omega_\text{cen}\mathcal{L}_\text{cen} + \omega_\text{box}\mathcal{L}_\text{box} + \omega_\text{ss}\mathcal{L}_\text{ss}
\end{equation}
where $\omega_\text{cls}$, $\omega_\text{cen}$, $\omega_\text{box}$ are set to 1, 0.1, 1 by default, and $\omega_\text{ss}$ is set to 0.3 (flip/rotate) or 0.02 (scale).

\subsection{Label Assignment}\label{sec:tech}
\vspace{-2pt}
Currently, available detectors largely rely on FPN (Feature Pyramid Network) \cite{Lin2017Feature}. For example, Rotated FCOS \cite{Tian2019FCOS} usually uses five feature layers, with large and small objects assigned to different ones. While point annotations do not provide any size information, they do not apply to such an FPN-based assignment strategy.

In terms of YOLOF \cite{chen2021yolof} that barely uses a one-level feature, it uses five preset anchors with sizes 32, 64, 128, 256, and 512. Therefore its anchor assignment strategy is also incompatible with point annotations.

To mitigate this problem, we propose a classification-score-based assignment rule. Instead of assigning ground-truths to the anchor with the highest IoU, we assign them (including both labeled points and synthetic boxes) to the one that produces the highest classification score.

Specifically, YOLOF is used as the backbone detector. The anchor stride is 16 and all five anchors are set to a fixed size ($64 \times 64$ for the DOTA dataset and $128 \times 128$ for others). Then the matching scores between anchors and ground-truths can be calculated as:
\begin{equation}
\setlength{\abovedisplayskip}{6pt}
\setlength{\belowdisplayskip}{6pt}
\mathit{ score}= \begin{cases}
0, \quad\quad L_1\left ( xy_\text{pred},xy_\text{gt} \right )>32\\
L_\text{cls}\left(c_\text{pred}, c_\text{gt} \right ) , \quad\mathit{otherwise}
\end{cases} 
\end{equation}
where $xy_\text{pred}$ and $xy_\text{gt}$ are the center coordinates of predicted boxes and ground-truths; $c_\text{pred}$ and $c_\text{gt}$ are the predicted classification scores and ground-truth labels. 

Afterward, following the setting of YOLOF, we use K-nearest to find four positive anchors with the highest scores for each ground truth. The improvement of the above design is verified in the ablation study (see Sec. \ref{sec:ablation}).

\subsection{Inference Phase}\label{sec:infer}
\vspace{-2pt}
Although using a synthetic pattern generation module and two branches in training (see Fig.~\ref{fig:point2rbox}), Point2RBox does not require these operations in inference. Due to the parameter sharing of the two branches, the inference only involves the forward propagation of the backbone, the classification head, and the regression head. Thus, it has a similar inference speed compared to the backbone detector that it is based on.

\section{Experiments}\label{sec:experiments}
\vspace{-4pt}
Experiments are carried out on NVIDIA RTX4090/A100 GPUs using PyTorch 1.13.1 \cite{Paszke2019PyTorch} and the rotation detection tool kits: MMRotate 1.0.0 \cite{Zhou2022MMRotate}. All the experiments follow the same hyper-parameters (learning rate, batch size, optimizer, etc.).


\begin{table*}[t]
\fontsize{8.5pt}{11pt}\selectfont
\setlength{\tabcolsep}{1.46mm}
\setlength{\aboverulesep}{0.4ex}
\setlength{\belowrulesep}{0.4ex}
\setlength{\abovecaptionskip}{1.5mm}
\centering
\begin{tabular}{l|ccccccccccccccc|c}
\toprule
\textbf{Methods}  & \textbf{PL}    & \textbf{BD}    & \textbf{BR}    & \textbf{GTF}   & \textbf{SV}    & \textbf{LV}    & \textbf{SH}    & \textbf{TC}    & \textbf{BC}    & \textbf{ST}    & \textbf{SBF}   & \textbf{RA}    & \textbf{HA}    & \textbf{SP}    & \textbf{HC}    & \textbf{AP}$_\text{50}$  \\ \hline
\multicolumn{17}{l}{\textbf{RBox-supervised}}                                                                                                                \\ \hline
RepPoints (2019) \cite{Yang2019Reppoints}            & 86.7  & 81.1  & 41.6  & 62.0  & 76.2  & 56.3  & 75.7  & 90.7  & 80.8  & 85.3  & 63.3 & 66.6  & 59.1  & 67.6  & 33.7  & 68.45 \\ 
RetinaNet (2017) \cite{Lin2017Focal}            & 88.2  & 77.0  & 45.0  & 69.4  & 71.5  & 59.0  & 74.5  & 90.8  & 84.9  & 79.3  & 57.3 & 64.7  & 62.7  & 66.5  & 39.6  & 68.69 \\
GWD (2021) \cite{Yang2021Rethinking}                  & 89.3  & 75.4  & 47.8  & 61.9  & 79.5  & 73.8  & 86.1  & 90.9  & 84.5  & 79.4  & 55.9 & 59.7  & 63.2  & 71.0  & 45.4  & 71.66 \\
FCOS (2019) \cite{Tian2019FCOS}                 & 89.1  & 76.9  & 50.1  & 63.2  & 79.8  & 79.8  & 87.1  & 90.4  & 80.8  & 84.6  & 59.7 & 66.3  & 65.8  & 71.3  & 41.7  & 72.44 \\
S$^2$A-Net (2022) \cite{Han2022Align}       & 89.2  & 83.0  & 52.5  & 74.6  & 78.8  & 79.2  & 87.5  & 90.9  & 84.9  & 84.8  & 61.9 & 68.0  & 70.7  & 71.4  & 59.8  & \textbf{75.81} \\
YOLOF-A5 (2021) \cite{chen2021yolof} $^1$               & 86.9  & 71.7  & 44.9  & 62.7  & 69.9  & 62.0  & 74.1  & 90.9  & 78.4  & 72.9  & 47.1 & 60.7  & 61.3  & 65.3  & 50.2  & 66.54 \\
YOLOF-A1 (2021) \cite{chen2021yolof} $^1$               & 68.7  & 52.3  & 25.2  & 36.8  & 50.6  & 47.8  & 58.7  & 81.6  & 64.5  & 65.4  & 16.9  & 56.0  & 30.0  & 51.4  & 43.5  & 49.94 \\ \hline
\multicolumn{17}{l}{\textbf{HBox-to-RBox}}                                                                                                                    \\ \hline
Sun et al. (2021) \cite{sun2021oriented} & 51.5  & 38.7  & 16.1  & 36.8  & 29.8  & 19.2  & 23.4  & 83.9  & 50.6  & 80.0  & 18.9 & 50.2  & 25.6  & 28.7  & 25.5  & 38.60 \\
BoxInst-RBox (2021) \cite{tian2021boxinst} $^2$         & 68.4 & 40.8 & 33.1 & 32.3 & 46.9 & 55.4 & 56.6 & 79.5 & 66.8 & 82.1 & 41.2 & 52.8 & 52.8 & 65.0 & 30.0 & 53.59 \\
H2RBox (2023) \cite{yang2023h2rbox}               & 88.5  & 73.5  & 40.8  & 56.9  & 77.5  & 65.4  & 77.9  & 90.9  & 83.2  & 85.3  & 55.3 & 62.9  & 52.4  & 63.6  & 43.3  & 67.82 \\
H2RBox-v2 (2023) \cite{yu2023h2rboxv2}           & 89.0 & 74.4 & 50.0 & 60.5 & 79.8 & 75.3 & 86.9 & 90.9 & 85.1 & 85.0 & 59.2 & 63.2 & 65.2 & 70.5 & 49.7 & \textbf{72.31} \\ \hline
\multicolumn{17}{l}{\textbf{Point-to-RBox}}                                                                                                                   \\ \hline
\footnotesize Point2Mask-RBox (2023) \cite{li2023point2mask} $^2$   & 4.0 & 23.1 & 3.8 & 1.3 & 15.1 & 1.0 & 3.3 & 19.0 & 1.0 & 29.1 & 0.0 & 9.5 & 7.4 & 21.1 & 7.1 & 9.72 \\
\scriptsize P2BNet+H2RBox (2023) \cite{chen2022pointtobox,yang2023h2rbox} $^3$ & 24.7 & 35.9 & 7.1 & 27.9 & 3.3 & 12.1 & 17.5 & 17.5 & 0.8 & 34.0 & 6.3 & 49.6 & 11.6 & 27.2 & 18.8 & 19.63 \\
\scriptsize P2BNet+H2RBox-v2 (2023) \cite{chen2022pointtobox,yu2023h2rboxv2} $^3$   & 11.0 & 44.8 & 14.9 & 15.4 & 36.8 & 16.7 & 27.8 & 12.1 & 1.8 & 31.2 & 3.4 & 50.6 & 12.6 & 36.7 & 12.5 & 21.87\\
\rowcolor{gray!20} Point2RBox-RC (ours) $^4$ & 62.9 & 64.3 & 14.4 & 35.0 & 28.2 & 38.9 & 33.3 & 25.2 & 2.2  & 44.5 & 3.4  & 48.1 & 25.9 & 45.0 & 22.6 & 34.07 \\
\rowcolor{gray!20} Point2RBox-SK (ours) $^4$ & 53.3 & 63.9 & 3.7  & 50.9 & 40.0 & 39.2 & 45.7 & 76.7 & 10.5 & 56.1 & 5.4  & 49.5 & 24.2 & 51.2 & 33.8 & 40.27 \\
\rowcolor{gray!20} Point2RBox-SK (CSPNeXt) $^4$ & 63.9 & 49.9 & 11.7  & 48.4 & 42.2 & 43.4 & 51.5 & 90.6 & 3.0 & 53.3 & 3.0  & 45.1 & 20.5 & 50.9 & 38.2 & 41.05 \\ 
\rowcolor{gray!20} \footnotesize Point2RBox-SK (two-stage) $^{4,5}$ & 66.4 & 59.5 & 5.2 & 52.6 & 54.1 & 53.9 & 57.3 & 90.8 & 3.2 & 57.8 & 6.1  & 47.4 & 22.9 & 55.7 & 40.5 & \textbf{44.90} \\ \bottomrule
\specialrule{0pt}{2pt}{0pt}
\multicolumn{17}{l}{$^1$ A5: Using five anchor sizes (16, 32, 64, 128, 256). A1: Using single anchor size (64). Our Point2RBox is based on YOLOF-A1.} \\
\multicolumn{17}{l}{$^2$ -RBox: The minimum rectangle operation is performed on the output Mask to obtain the RBox.} \\
\multicolumn{17}{l}{$^3$ Using P2BNet (2022) \cite{chen2022pointtobox} for Point-to-HBox and then H2RBox-v2 (2023) \cite{yu2023h2rboxv2} for HBox-to-RBox.} \\
\multicolumn{17}{l}{$^4$ RC: Using rectangles and circles with curve textures as basic patterns. SK: Using one sketch pattern for each category as basic patterns.} \\
\multicolumn{17}{l}{$^5$ two-stage: Point2RBox-SK (A1) is trained to generate pseudo RBoxes, which enable the anchor-based assignment in the second stage (A5).} \\ \specialrule{0pt}{1pt}{0pt}
\bottomrule
\end{tabular}
\caption{Detection performance of each category on the DOTA-v1.0 and the mean AP$_\text{50}$ of all categories.}
\label{tab:exp_dota}
\vspace{-6pt}
\end{table*}

\begin{table}[t]
\fontsize{8.5pt}{11pt}\selectfont
\setlength{\tabcolsep}{2.8mm}
\setlength{\aboverulesep}{0.4ex}
\setlength{\belowrulesep}{0.4ex}
\setlength{\abovecaptionskip}{1.5mm}
\centering
\begin{tabular}{l|c|c}
\toprule
\textbf{Methods}     & ~~\textbf{DIOR}~~  & ~~\textbf{HRSC}~~  \\
\hline
\multicolumn{3}{l}{\textbf{RBox-supervised}} \\
\hline
RetinaNet (2017) \cite{Lin2017Focal} & 54.60 & 84.49 \\
GWD (2021) \cite{Yang2021Rethinking} & 57.80 & 86.67 \\
FCOS (2019) \cite{Tian2019FCOS} & \textbf{58.60} & 88.99 \\
YOLOF-A5 (2021) \cite{chen2021yolof} $^1$ & 48.01 & \textbf{89.44} \\
YOLOF-A1 (2021) \cite{chen2021yolof} $^2$ & 37.51 & 81.14 \\
\hline
\multicolumn{3}{l}{\textbf{HBox-to-RBox}}     \\
\hline
H2RBox (2023) \cite{yang2023h2rbox} & \textbf{57.00} & 7.03  \\
KCR (2023) \cite{zhu2023knowledge} & -     & 79.10 \\
H2RBox-v2 (2023) \cite{yu2023h2rboxv2} & 56.92 & \textbf{89.66}  \\
\hline
\multicolumn{3}{l}{\textbf{Point-to-RBox}}    \\
\hline
Point2Mask-RBox (2023) \cite{li2023point2mask} & 13.77 & 29.95 \\
P2BNet+H2RBox (2023) \cite{chen2022pointtobox,yang2023h2rbox}     & 22.59 & - \\
P2BNet+H2RBox-v2 (2023) \cite{chen2022pointtobox,yu2023h2rboxv2}     & 23.61 & 14.60 \\
\rowcolor{gray!20} Point2RBox-RC (ours) & 24.66 & 78.77 \\
\rowcolor{gray!20} Point2RBox-SK (ours) & 27.34 & 79.40 \\
\rowcolor{gray!20} Point2RBox-SK (CSPNeXt) & \textbf{27.62}  & \textbf{80.01} \\
\bottomrule
\specialrule{0pt}{2pt}{0pt}
\multicolumn{3}{l}{$^1$ A5: Using five anchor sizes (16, 32, 64, 128, 256).}\\
\multicolumn{3}{l}{$^2$ A1: Using single anchor size (128).} \\
\specialrule{0pt}{1pt}{0pt}
\bottomrule
\end{tabular}
\caption{AP$_{50}$ on the DIOR and the HRSC datasets.}
\label{tab:exp_hrsc}
\vspace{-8pt}
\end{table}

\subsection{Settings and Datasets}
\vspace{-2pt}
We use the FPN-free detector YOLOF \cite{chen2021yolof} with a fixed anchor size (64 for DOTA and 128 for DIOR/HRSC) as the baseline method for developing our Point2RBox. Such a choice is mainly upon the fact that positive samples (annotated with points) cannot be assigned to different FPN layers or different anchors based on their annotated sizes. 

Average precision (AP) is adopted as the primary metric to compare our methods with existing alternatives. Unless otherwise specified, all the listed models are configured based on ResNet50 \cite{He2016Deep} backbone. All models are trained with AdamW \cite{loshchilov2018decoupled}, with an initial learning rate of 5e-5 and a mini-batch size of 4. Besides, we adopt a learning rate warm-up for 500 iterations, and the learning rate is divided by ten at each decay step. 
“1x” and “6x” schedules indicate 12 and 72 epochs for training, and “RR” denotes random rotation augmentation. “1x” is used for the DOTA and DIOR datasets and “6x+RR” for HRSC. Random flipping and shifting are always adopted by default. For a fair comparison, all results are evaluated without multi-scale technique \cite{Zhou2022MMRotate}.

\textbf{DOTA \cite{Xia2018DOTA}.} DOTA-v1.0 contains 2,806 aerial images---1,411 for training, 937 for validation, and 458 for testing, as annotated using 15 categories with 188,282 instances in total. 
The categories are defined as: Plane (PL), Baseball Diamond (BD), Bridge (BR), Ground Track Field (GTF), Small Vehicle (SV), Large Vehicle (LV), Ship (SH), Tennis Court (TC), Basketball Court (BC), Storage Tank (ST), Soccer-Ball Field (SBF), Roundabout (RA), Harbor (HA), Swimming Pool (SP), and Helicopter (HC). 
We follow the default preprocessing in MMRotate: The high-resolution images are split into 1,024 $\times$ 1,024 patches with an overlap of 200 pixels for training, and the detection results of all patches are merged to evaluate the performance.

\textbf{DIOR \cite{cheng2022anchor}.} DIOR-RBox is an aerial image dataset re-annotated with RBoxes based on its original HBox-annotated version DIOR \cite{li2020object}. There are 23,463 images and 190,288 instances with 20 classes. DIOR-RBox has a high variation in object size with high intra‐class diversity.

\textbf{HRSC \cite{Liu2017HRSC}.} It contains ship instances both on the sea and inshore, with arbitrary orientations. The training, validation, and testing set includes 436, 181, and 444 images, respectively. With preprocessing by MMRotate, images are scaled to 800 $\times$ 800 for training/testing.

\subsection{Main Results}\label{sec:mainres}
\vspace{-2pt}


\textbf{DOTA-v1.0.} Table \ref{tab:exp_dota} shows that Point2RBox outperforms currently available two-stage solution -- Point-to-HBox-to-RBox, even if the Point-to-HBox-to-RBox is powered by the state-of-the-art methods P2BNet \cite{chen2022pointtobox} (Point-to-HBox) and H2RBox-v2 \cite{yu2023h2rboxv2} (HBox-to-RBox). 

Since point annotations cannot be assigned to different FPN layers or different anchors by size, our Point2RBox is based on the FPN-free method YOLOF-A1 \cite{chen2021yolof}. The performance gap between Point2RBox and the RBox-supervised YOLOF-A1 baseline is 9.67\% (40.27\% vs. 49.94\%), proving the effectiveness of combining knowledge from synthetic patterns for point-supervised oriented object detection.

Barely using synthetic patterns (i.e. SetRC), Point2RBox achieves AP$_\text{50}$ of 34.07\%. While one sketch pattern for each category is cooperated (only fifteen patterns in total, i.e. SetSK), the performance is further boosted to 40.27\%. By utilizing a stronger backbone CSPNeXt-l \cite{lyu2022rtmdet}, Point2RBox obtains 41.05\% on DOTA-v1.0.

In the last row of Table \ref{tab:exp_dota}, we additionally explore a two-stage training pipeline. In the first stage, Point2RBox-SK (based on YOLOF-A1) is trained to generate pseudo RBoxes labels. These pseudo labels enable the anchor-based assignment in the second stage (based on YOLOF-A5), which improves the performance to 44.90\%.

\textbf{DIOR.} Table \ref{tab:exp_hrsc} shows that Point2RBox achieves AP$_\text{50}$ of 24.66\% and 27.34\% in RC and SK settings respectively. 

Compared with the state-of-the-art Point-to-HBox-to-RBox solution (i.e. P2BNet \cite{chen2022pointtobox} + H2RBox-v2 \cite{yu2023h2rboxv2}), our method uses a light-weight end-to-end paradigm, yet obtains a competitive performance (27.34\% vs. 22.30\%). 

\textbf{HRSC.} Table \ref{tab:exp_hrsc} shows that Point2RBox achieves 78.77\% in the SetRC and 79.40\% in the SetSK. Notably, HRSC only has one category, so we also use only one pattern cropped from the first training image as the basic pattern for the SK setting. A previous work KCR (2023) \cite{zhu2023knowledge} combines knowledge from RBox-annotated DOTA dataset to achieve HBox-to-RBox on HRSC. Compared with that, our SetSK method, under a more challenging Point-to-RBox setting, outperforms KCR by 0.3\% (79.40\% vs. 79.10\%). 

Compared to RBox-supervised counterpart YOLOF-A1 \cite{chen2021yolof}, the performance gap on HRSC is only 1.74\% (79.40\% vs. 81.14\%). Finally, Point2RBox obtains 80.01\% on HRSC by further utilizing a stronger CSPNeXt-l \cite{lyu2022rtmdet} backbone.

\textbf{Computational cost.} Point2RBox is light-weight in three folds: \textbf{1)} end-to-end; \textbf{2)} fewer parameters; \textbf{3)} anchor/FPN-free. Specifically, training Point2RBox-SK on DOTA takes about 5 hours and the inference speed is about 112 fps on our Intel i9-14900 + NVIDIA RTX4090 hardware.

\subsection{Ablation Studies}\label{sec:ablation}
\vspace{-2pt}
Several ablation studies are performed on Point2RBox to evaluate the impact of each proposed module.

\textbf{Basic pattern settings.} Table \ref{tab:abl_ss} studies the impact of different strategies to obtain basic patterns for synthetic generation. The ``Shape'' column indicates using rectangles and circles with white filling and black edges; the ``Curve'' column indicates adding curve textures on the rectangles and circles; the ``Sketch'' column indicates using one sketch pattern for each category (see Sec. \ref{sec:skc}). The three rows in Table \ref{tab:abl_ss} correspond to the three cases displayed in Fig. \ref{fig:settings}

Even if using simplest rectangles and circles, we show the knowledge can be well combined, resulting in AP$_\text{50}$ of 75.01\% on HRSC. When curve textures are added (i.e. SetRC), the performance is further boosted to 78.77\%, proving the effectiveness of using curve textures. Sketch patterns (i.e. SetSK) provide more accurate semantic boundaries, leading to an improvement of 6.20\%/2.68\%/0.63\% on DOTA/DIOR/HRSC compared to the SetRC.

\begin{table}[t]
\fontsize{8.5pt}{11pt}\selectfont
\setlength{\tabcolsep}{3.4mm}
\setlength{\aboverulesep}{0.4ex}
\setlength{\belowrulesep}{0.4ex}
\setlength{\abovecaptionskip}{1.5mm}
\centering
\begin{tabular}{c|ccc|c}
\toprule
\textbf{Datasets}               & \textbf{Shape} & \textbf{Curve} & \textbf{Sketch} & ~~~\textbf{AP}$_\text{50}$~~~  \\ \hline
\multirow{3}{*}{DOTA \cite{Xia2018DOTA}} & \checkmark     &       &        &   29.72   \\
                      &   \checkmark   & \checkmark     &        & 34.07 \\
                      &  \cellcolor{gray!20}     &    \cellcolor{gray!20}   & \cellcolor{gray!20} \checkmark~      & \cellcolor{gray!20} \textbf{40.27}~ \\ \hline
\multirow{3}{*}{DIOR \cite{cheng2022anchor}} & \checkmark     &       &        & 22.23  \\
                      &   \checkmark   & \checkmark     &        & 24.66 \\
                      &   \cellcolor{gray!20}    &   \cellcolor{gray!20}    & \cellcolor{gray!20} \checkmark~      & \cellcolor{gray!20} \textbf{27.34}~ \\ \hline
\multirow{3}{*}{HRSC \cite{Liu2017HRSC}} & \checkmark     &       &        & 75.01 \\
                      &   \checkmark   & \checkmark     &        & 78.77 \\
                      &   \cellcolor{gray!20}    &    \cellcolor{gray!20}   & \cellcolor{gray!20} \checkmark~      & \cellcolor{gray!20} \textbf{79.40}~ \\ \bottomrule
\end{tabular}\caption{Ablation with different basic pattern settings.}
\label{tab:abl_pattern}
\vspace{-8pt}
\end{table}

\textbf{Transform self-supervision.} Table \ref{tab:abl_ss} studies the impact of using different transformations in the transform self-supervision (see Sec. \ref{sec:tss}). Rotation and flipping are adopted together since they both act on the angle. According to the results, when self-supervised by rotated and flipped views, the performance is boosted by 4.29\% on average, whereas incorporating scaling gains an additional 1.82\% improvement. The results prove that the transform self-supervision plays a crucial role in the proposed training paradigm.

\textbf{Annotation inaccuracy.} Table \ref{tab:abl_noise} offsets the coordinates of annotated points by a noise from the uniform distribution $\left[-\sigma H, +\sigma H \right ]$, where $H$ is the height of objects. Such a setting is designed to simulate the noise in the real annotation. When $\sigma = 10\%$, AP$_{50}$ on the DIOR dataset is even improved to 30.82\%. For some categories, e.g. basketball court with a circle in the center, the network sometimes takes the size of the central circle as the size of the basketball court. Adding noise alleviates this issue, which may explain the performance improvement. When $\sigma = 20\%$, the AP$_{50}$ of Point2RBox drops by only 1.03\% on average, which demonstrates the robustness of our method.

\textbf{Label assignment.} With a fixed anchor size, AP$_{50}$ of using one/three/five anchors is 37.15\%/39.48\%/40.27\% on DOTA, which proves that using multiple anchors of the same size (assigned based on classification score, see Sec. \ref{sec:tech}) can improve performance to some extent.

\textbf{Knowledge combination.} Our method's effectiveness lies in spreading features near each labeled point to the generated patterns, which narrows the gap between the synthetic and real data. With this key recolor step removed (i.e. directly pasting augmented patterns like copy-paste), the AP$_{50}$ is much lower (40.27\% vs. 28.72\%) on DOTA (SetSK).

\begin{table}[t]
\fontsize{8.5pt}{11pt}\selectfont
\setlength{\tabcolsep}{3.76mm}
\setlength{\aboverulesep}{0.4ex}
\setlength{\belowrulesep}{0.4ex}
\setlength{\abovecaptionskip}{1.5mm}
\centering
\begin{tabular}{c|ccc|c}
\toprule
\textbf{Datasets}               & \textbf{Rotate} & \textbf{Flip} & \textbf{Scale} & ~~~\textbf{AP}$_\text{50}$~~~  \\ \hline
\multirow{3}{*}{DIOR \cite{cheng2022anchor}} &      &       &        &   21.34   \\
                      &   \checkmark   & \checkmark     &        & 24.97 \\
                      &   \cellcolor{gray!20} \checkmark~   &  \cellcolor{gray!20} \checkmark~   & \cellcolor{gray!20} \checkmark~      & \cellcolor{gray!20} \textbf{27.34}~ \\ \hline
\multirow{3}{*}{HRSC \cite{Liu2017HRSC}} &      &       &        & 73.18 \\
                      &   \checkmark   & \checkmark     &        & 78.13 \\
                      &   \cellcolor{gray!20} \checkmark~   &  \cellcolor{gray!20} \checkmark~   & \cellcolor{gray!20} \checkmark~      & \cellcolor{gray!20} \textbf{79.40}~ \\ \bottomrule
\end{tabular}\caption{Ablation with the transform self-supervision.}
\label{tab:abl_ss}
\vspace{-2pt}
\end{table}

\begin{table}[t]
\fontsize{8.5pt}{11pt}\selectfont
\setlength{\tabcolsep}{4.6mm}
\setlength{\aboverulesep}{0.4ex}
\setlength{\belowrulesep}{0.4ex}
\setlength{\abovecaptionskip}{1.5mm}
\centering
\begin{tabular}{c|c|c|c}
\toprule
\textbf{Datasets}               & ~$\sigma$ = 0\%~ & $\sigma$ = 10\% & $\sigma$ = 20\% \\ \hline
DOTA \cite{Xia2018DOTA} & \cellcolor{gray!20} \textbf{40.27}~ & 39.60 & 38.42 \\ 
DIOR \cite{cheng2022anchor} & 27.34 & \cellcolor{gray!20} \textbf{30.82}~ & 27.22 \\ 
HRSC \cite{Liu2017HRSC} & \cellcolor{gray!20} \textbf{79.40}~ & 78.81 & 78.28 \\ \bottomrule
\end{tabular}\caption{Ablation with the inaccuracy in point annotations.}
\label{tab:abl_noise}
\vspace{-8pt}
\end{table}

\section{Conclusion}\label{sec:conclusion}
\vspace{-4pt}
This paper has presented Point2RBox, a weakly-supervised oriented object detector that learns from the point annotation. It adopts an end-to-end paradigm to directly obtain the RBox prediction through knowledge combination from synthetic visual patterns, which has the advantage of being concise and cost-efficient over the two-stage alternatives (e.g. Point-to-HBox-to-RBox or generating RBox pseudo labels from the Mask). Supplemented with the transform self-supervision, the performance is further improved. 

Experiments are carried out with the following observations: 
\textbf{1)} Point2RBox achieves point-supervised oriented object detection in an end-to-end training manner. Upon that, we show the knowledge from synthetic patterns can be combined to estimate the size and angle of real objects. 
\textbf{2)} Compared with KCR \cite{zhu2023knowledge} that combines knowledge for HBox-supervised setting, our method outperforms KCR by 0.3\% (HRSC: 79.40\% vs. 79.10\%) under a more challenging point-supervised setting. 
\textbf{3)} Our method outperforms the state-of-the-art alternative (i.e. P2BNet \cite{chen2022pointtobox} + H2RBox-v2 \cite{yu2023h2rboxv2}) by a large margin based on the same ResNet50 backbone (DOTA/DIOR/HRSC: 40.27\%/27.34\%/79.40\% vs. 21.87\%/22.30\%/14.60\%). With CSPNeXt-l \cite{lyu2022rtmdet} backbone, the performance reaches 41.05\%/27.62\%/80.01\%, proving the super effectiveness of the proposed method. 

\textbf{Limitations. 1)} Point2RBox gives a bad performance on some categories (i.e. BR/BC/SBF), mainly due to their non-unique boundaries. For instance, the central circle of the basketball court is often mistaken as the boundary. \textbf{2)} Point2RBox can only be built upon FPN-free detectors (e.g. YOLOF-A1) since points cannot be assigned to different FPN layers/anchors based on their sizes.

{
    \small
    \bibliographystyle{ieeenat_fullname}
    \bibliography{main}
}


\end{document}